%% file: ICCV.tex
\documentclass[10pt,twocolumn,letterpaper]{article}

\usepackage{iccv}
\usepackage{times}
\usepackage{epsfig}
\usepackage{graphicx}
\usepackage{amsmath}
\usepackage{amssymb}
\usepackage{multirow}
\usepackage{mathrsfs}
\usepackage{tabulary}
\usepackage[dvipsnames]{xcolor}
\usepackage{colortbl}
\definecolor{mygray}{gray}{0.6}
\definecolor{mygray-bg}{gray}{0.9}
\usepackage{graphicx}
\usepackage{amsmath}
\usepackage{amssymb}
\usepackage{booktabs}
\usepackage{multirow}
\usepackage{pifont}
\usepackage{booktabs}
\usepackage{epsfig}
\usepackage{booktabs,multirow}
\usepackage{graphics}
\usepackage{threeparttable}
\usepackage{color}
\usepackage[normalem]{ulem}
\usepackage{multirow}
\usepackage{float}
\usepackage{amsfonts}
\usepackage{bm}
\usepackage{subfig}
\usepackage{enumitem}
\usepackage[numbers]{natbib}
\usepackage{array}
\usepackage{colortbl}
\usepackage[export]{adjustbox}
\usepackage{bm}
\usepackage{amsmath}
\usepackage{gensymb}
\usepackage{wrapfig}

\newcolumntype{I}{!{\vrule width 1pt}}

\makeatletter
\newcommand{\thickhline}{%
    \noalign {\ifnum 0=`}\fi \hrule height 1pt
    \futurelet \reserved@a \@xhline
}
\makeatother

\usepackage{caption}
\iccvfinalcopy
\usepackage[pagebackref=true,breaklinks=true,letterpaper=true,colorlinks,bookmarks=false]{hyperref}

\newcommand{\app}{\raise.17ex\hbox{$\scriptstyle\sim$}}

\newcolumntype{x}[1]{>{\centering\arraybackslash}p{#1pt}}
\newlength\savewidth\newcommand\shline{\noalign{\global\savewidth\arrayrulewidth\global\arrayrulewidth 1pt}\hline\noalign{\global\arrayrulewidth\savewidth}}

\def\iccvPaperID{5514} 

\ificcvfinal\fi

\begin{document}

\title{Omnidirectional Information Gathering for\\Knowledge Transfer-based Audio-Visual Navigation}
\author{Jinyu Chen$^{1}$, Wenguan Wang$^{2*}$, Si Liu$^{1}$\thanks{Corresponding author: \textit{Wenguan Wang, Si Liu}.}~, Hongsheng Li$^{3}$, Yi Yang$^{2}$\\
\small{$^1$ Institute of Artificial Intelligence, Beihang University}  
\small{$^2$ ReLER, CCAI, Zhejiang University}  
\small{$^3$ The Chinese University of Hong Kong} \\
\small\url{https://github.com/chenjinyubuaa/ORAN}
}
\maketitle
\ificcvfinal\thispagestyle{empty}\fi


\maketitle
\begin{abstract}
Audio-visual navigation is an audio-targeted wayfinding task where a robot agent is entailed to travel a never-before-seen 3D environment towards the sounding source. In this article, we present \texttt{ORAN}, an \underline{o}mnidi\underline{r}ectional \underline{a}udio-visual \underline{n}avigator based on cross-task navigation skill transfer. In particular, \texttt{ORAN} sharpens its two basic abilities for a such challenging task, namely wayfinding and audio-visual information gathering. First, \texttt{ORAN} is trained with a confidence-aware cross-task policy distillation (CCPD) strategy. CCPD transfers the fundamental, point-to-point wayfinding skill that is well trained on the large-scale PointGoal task to \texttt{ORAN}, so as to help \texttt{ORAN} to better master audio-visual navigation with far fewer training samples. To improve the efficiency of knowledge transfer and address the domain gap, CCPD is made to be adaptive to the decision confidence of the teacher policy. Second, \texttt{ORAN} is equipped with an omnidirectional information gathering (OIG) mechanism, \ie,  gleaning visual-acoustic observations from different directions before decision-making. As a result, \texttt{ORAN} yields more robust navigation behaviour. Taking CCPD and OIG together, \texttt{ORAN} significantly outperforms previous competitors. After the model ensemble, we got 1st in  Soundspaces Challenge 2022, improving SPL and SR by 53\% and 35\% relatively.

\end{abstract}
\begin{figure}
\vspace{5mm}
   \begin{center}
      \includegraphics[width=1\linewidth]{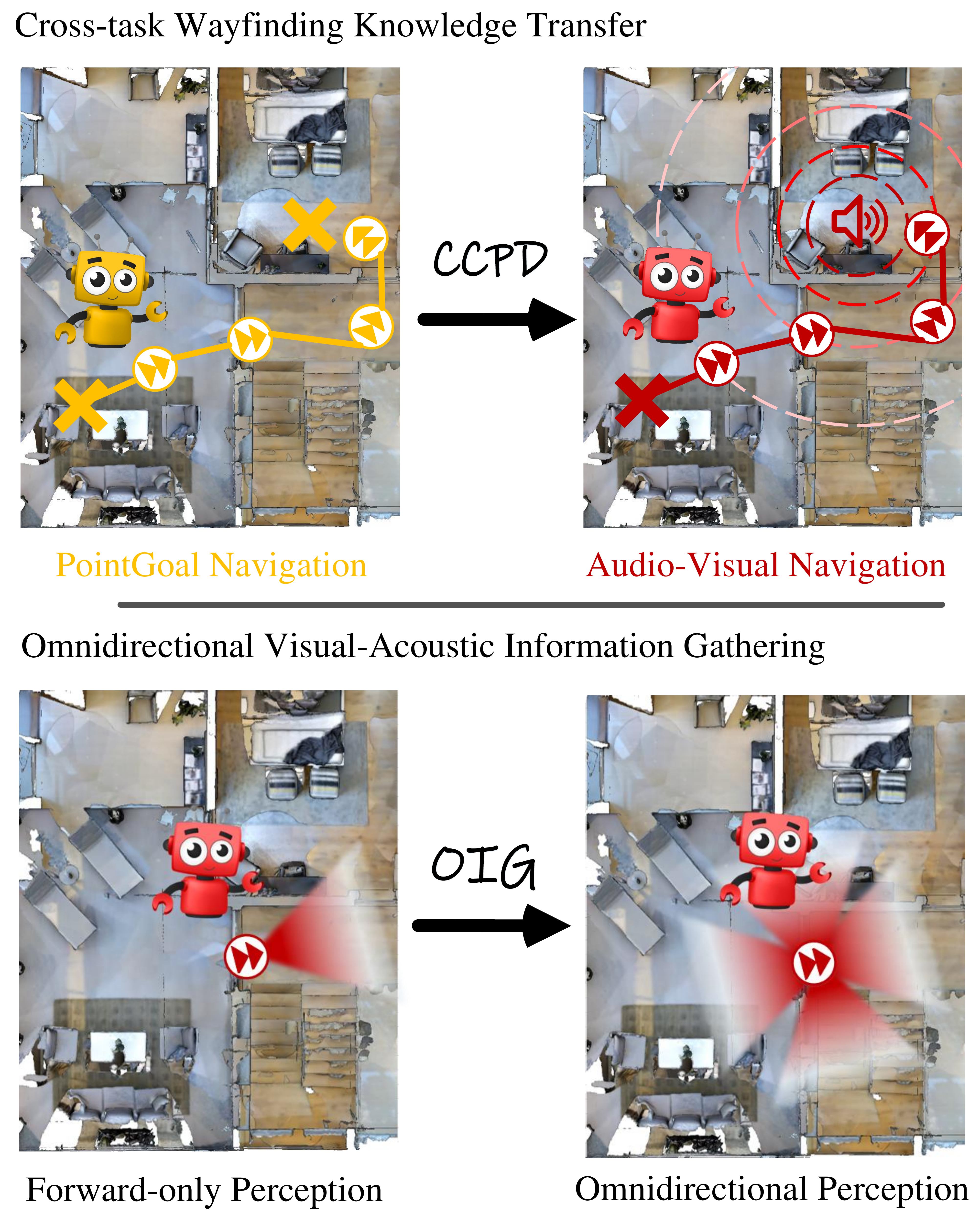}
   \end{center}\vspace{-4mm}
      \caption{Top: Based on CCPD, our \texttt{ORAN} transfers wayfinding knowledge well-trained on PointGoal task to audio-visual navigation. Bottom: Based on OIG, \texttt{ORAN} collects visual and acoustic information from different directions for robust decision-making.
      }
   \label{fig:fig1}
\end{figure}
\section{Introduction}                                        
Developing intelligent autonomous wayfinding agents that can robustly navigate in unexplored environments to reach target locations is one of the most classic and fundamental tasks in robotics and is widely viewed as a critical building block of embodied AI. To simulate different real-world application scenarios of wayfinding agents, various navigation tasks are proposed, where the target goal is appointed by, for example, GPS coordinate~\cite{ppo, Georgakis_2022_CVPR,georgakis2022uncertainty}, semantic tag~\cite{ObjectGoal01, ObjectGoal02, ObjectGoal03, ObjectGoal04}, visual language instruction~\cite{fried2018speaker, qi2020reverie,anderson2018vision, Gao_2021_CVPR,an2022bevbert, qiao2023hop+, hong2011recurrent, zhu2021diagnosing, qiao2022hop, an2021neighbor,zhao2022target,wang2022counterfactual,wang2020active,wang2021structured,chen2022reinforced}, image photo~\cite{chen2021semantic,al2022zero,lyu2022improving,gupta2017cognitive}. Among these navigation tasks, in this article, we are particularly interested in audio-visual navigation (also named as AudioGoal)~\cite{soundspaces, AudioVisualNavigation00}, in which the navigation agent is entailed to find sound-emitting objects in visually-and-acoustically rich 3D environments. Audio plays an irreplaceable role here -- it reveals not only the properties of the target object but also the layout of the unexplored areas. For example, when we hear a sound, we can know what makes the such sound and locate the sound source even before we observe it.
Along this direction, existing AudioGoal solutions are typically reinforcement learning (RL) algorithms for direct audio-visual perception to low-level navigation action mapping~\cite{soundspaces, savn, YinfengICLR2022saavn}. 
Some methods further predict the next intermediate goal~\cite{setwaypoints} on a top-down topology map to improve the agent's long-term stability. 
Methods like~\cite{YinfengICLR2022saavn} instead consider the sound attacks to promote navigation robustness in complicated audio environments.


In this article, we develop \texttt{ORAN}, an  \underline{o}mnidi\underline{r}ectional \underline{a}udio-visual \underline{n}avigation agent based on cross-task wayfinding skill transfer. As shown in Figure~\ref{fig:fig1},  \texttt{ORAN} advances state-of-the-art technologies for AudioGoal in two aspects, namely wayfinding and visual-audio information gathering. Our technique innovations are born from two crucial insights. \textbf{First}, besides comprehending the received visual and audio signals, a successful AudioGoal agent needs to master some very basic wayfinding skills, such as precisely moving towards a short-term target, safely travelling without collision, and entering/leaving a room through the door. It is clear that these basic wayfinding skills are shared among different wayfinding tasks. Hence it is reasonable to assume that an AudioGoal agent can benefit from the knowledge of a high-performance navigator that is already well-trained on other navigation tasks, especially considering that the training samples of AudioGoal are relatively limited, while AudioGoal itself is a very challenging task. \textbf{Second}, it is natural for our human to turn around upon hearing a sound behind us, or turn our head to find out what we have heard~\cite{brown2013consciousness}. In sharp contrast, existing AudioGoal agents only make use of visual-audio information perceived forward during navigation decision-making. It is clear that there exists a huge gap between such a simple, forward-only perception regime between the omnidirectional decision-making mode (particularly for those top-down map-based AudioGoal agents~\cite{setwaypoints, AudioVisualNavigation01}). 

Our \texttt{ORAN} is elaborately designed to be fully aware of the aforementioned issues, so as to sharpen its wayfinding and visual-audio information-gathering abilities. \textbf{First}, \texttt{ORAN} is trained with a Confidence-aware Cross-task Policy Distillation (CCPD) strategy. CCPD allows \texttt{ORAN} to transfer the navigation knowledge learned on PointGoal to AudioGoal. Recently, PointGoal~\cite{habitat2019,habitat2021} has seen significant advance -- with millions of frames of experience and assistance of a GPS+Compass sensor, PointGoal agents can achieve nearly perfect navigation performance, given the coordinates of starting and target locations~\cite{wijmans2019dd,Partsey_2022_CVPR}. During the training of AudioGoal, we consider the behaviours of a well-trained PointGoal agent, queried with the same starting and target waypoint locations, as informative demonstrations for \texttt{ORAN}. The reuse of the navigation knowledge relieves \texttt{ORAN}'s burden of learning to both masters basic navigation skills and how to understand and plan with visual-audio perception.  Moreover, to further better overcome the domain gap between the two tasks and improve the efficiency of such cross-task navigation knowledge transfer, we render larger weights to those more confident steps of the PointGoal policy during policy distillation. \textbf{Second}, \texttt{ORAN} is empowered with an omnidirectional information gathering (OIG) ability. OIG enables \texttt{ORAN} to make use of acoustic-visual information collected from different directions, instead of the only direction it is facing, to support its omnidirectional navigation decision-making.

We experimentally demonstrate that combining CCPD and OIG together makes our \texttt{ORAN} a powerful AudioGoal navigator, which sets state-of-the-art performance on Soundspaces Challenge dataset~\cite{soundspaces}, \eg, $>$10\% absolute lifting in SPL on \textit{unhead} sets.  Moreover, after model assembling, \texttt{ORAN} yields further 12\% absolute promotion in SPL. The fused model won 1st on the Soundspaces challenge 2022~\cite{soundspaces} and improves SPL by 53\% and SR by 35\% relatively. 
\section{Related Work}

\noindent\textbf{Audio-visual Navigation.} 
The audio-visual Navigation task, also referred to as AudioGoal navigation involves the exploration of unfamiliar surroundings, utilizing both visual and auditory cues in order to navigate towards a designated sound source. Various platforms are developed for simulating indoor audio-visual environments. For instance, VAR~\cite{AudioVisualNavigation00} employs the AI2-THOR~\cite{ai2thor} platform as a foundation to construct a simulated audio-visual navigation environment. Soundspaces~\cite{soundspaces}, on the other hand, utilizes the Habitat~\cite{habitat2021,habitat2019} simulator to create a photorealistic indoor environment that includes acoustic simulation. Additionally, Soundspaces2.0~\cite{chen22soundspaces2} allows for the continuous visual and acoustic simulation of arbitrary mesh indoor data.

To address the AudioGoal task, the majority of studies employ RL for training deep learning policies.
For instance, AV-Nav\cite{soundspaces} employs the Proximal Policy Optimization (PPO) RL algorithm~\cite{ppo} to train an LSTM model that predicts low-level actions (\eg, turning left, moving forward) based on raw RGB/depth images and audio spectrograms at the current step.
SAVi~\cite{savn} incorporates a goal descriptor network for sound source direction prediction and a transformer for sequence modelling, aimed at addressing the audio-visual navigation problem with a temporary sound source. Inspired by the study in other acoustic area~\cite{1-sound-attack,2-sound-attack,3-sound-attack,4-sound-attack,5-sound-attack}, \cite{YinfengICLR2022saavn} proposes sound attackers as a means of enhancing the model's robustness in real-world scenarios. \cite{wang2022towards} introduces an agent capable of performing various navigational tasks, encompassing audio signals as well.
Additionally, the incorporation of occupancy maps into the model is introduced in the works of \cite{AudioVisualNavigation00,setwaypoints}. The agent predicts intermediate goals on the map and utilizes a graph-based shortest path algorithm to reach the intended position. The implementation of these techniques reduces decision-making times, thus enhancing the agent's long-term navigation performance. 
The objective of this paper is to enhance the dependability and efficacy for the occupancy-map-based AudioGoal model.

\vspace{1mm}
\noindent\textbf{PointGoal Navigation.} PointGoal navigation is a classic problem in the robotics area. The objective of the task is for an agent to traverse from a randomized point of origin to a designated target location within an unfamiliar environment using GPS and visual data.
The traditional way is to compose this task into several sub-tasks, \eg, mapping, planning and control. 
Recently, \cite{wijmans2019dd} found that this task can be solved with RL based end-to-end model. After training with a large number of samples, the RNN-based navigation policy can achieve near-perfect performance. 
\cite{ye2021auxiliary} proposes to apply the auxiliary losses to speed up the training process of the PointGoal policy. 
\cite{gordon2019splitnet} attempts to apply the visual encoder of the PointGoal policy to other visual navigation tasks.
\cite{Partsey_2022_CVPR} proposes to use the visual odometry module for PointGoal navigation in the presence of noise, which yields outstanding results.
In this paper, we utilize the basic navigation knowledge in the PointGoal policy to improve the ability of the AudioGoal policy.

\vspace{1mm}
\noindent\textbf{Policy Distillation.}
The knowledge distillation technique is widely studied in different deep learning area~\cite{cho2019efficacy,zhao2022decoupled,deng2022anomaly,xing2022selfmatch}, which aims to improve the performance of the student model by having it learn from the insights and knowledge acquired by the more complex and accurate teacher model.
In RL, transferring knowledge from teacher policies has attracted great research interests. 
In particular, a student policy is trained to match the state-dependent probability distribution over actions provided by the teacher, while the student policy is usually of lighter architecture for faster inference.
\cite{Online2019Marc} devised an online policy distillation paradigm that enables student policies to have better adversarial robustness.
Recently, \cite{czarnecki2019distilling} designed a teacher reward-based method for better leveraging imperfect teachers. 
\cite{qu2022importance} introduces frame importance analysis for the policy distillation training.
In~\cite{DueDis}, they propose a dual policy distillation, in which two learners extract knowledge from each other to enhance their learning. 
In this paper, we apply policy distillation to transfer knowledge from PointGoal to AudioGoal. So far, cross-task policy distillation has been less studied in the field of embodied navigation.

\section{Our Approach}
We propose \texttt{ORAN}, a powerful AudioGoal navigation agent built with two essential technique contributions, namely i) confidence-aware cross-task policy distillation (CCPD, \S\ref{sec:CCPD}), and ii) omnidirectional visual-audio information gathering (OIG, \S\ref{sec:OIG}), as shown in Figure~\ref{fig:pipline}. Before elaborating on the algorithm details, we first provide a overview of our task setup and basic architecture (\S\ref{problem+network}). 

\begin{figure}
   \begin{center}
      \includegraphics[width=\linewidth]{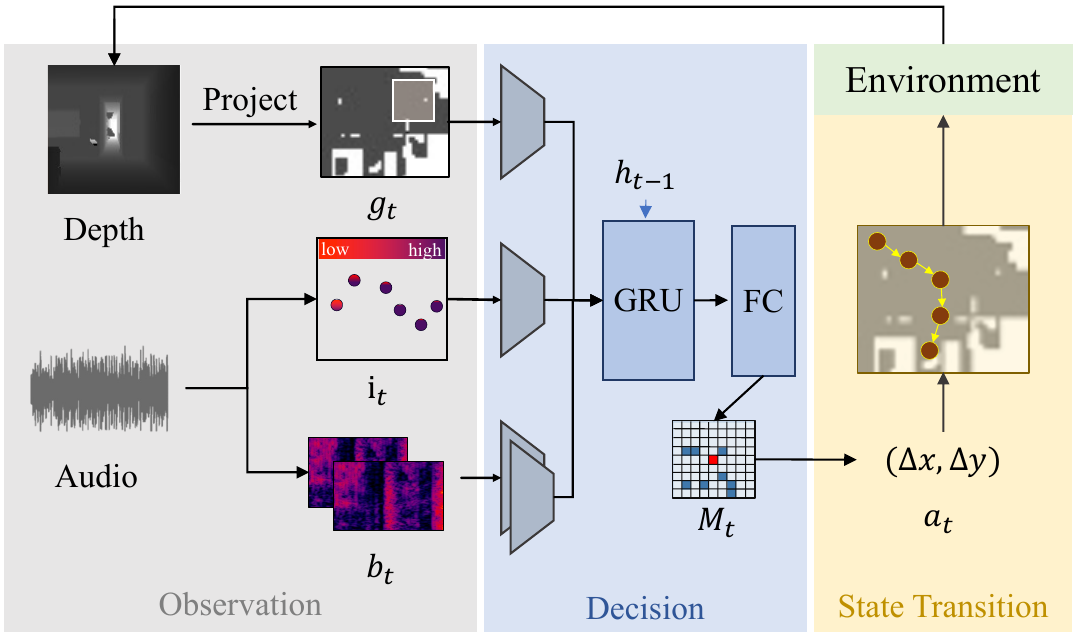}
   \end{center}
      \caption{ Illustration of the basic architecture. the navigation process can be expressed as the iterative execution of three steps, \ie, observation, decision and state transition.}
   \label{fig:network}
\end{figure}

\subsection{Problem Setup and Basic Architecture}\label{problem+network}
\begin{figure*}[t]
\vspace{2mm}
\begin{center}
          \includegraphics[width=\linewidth]{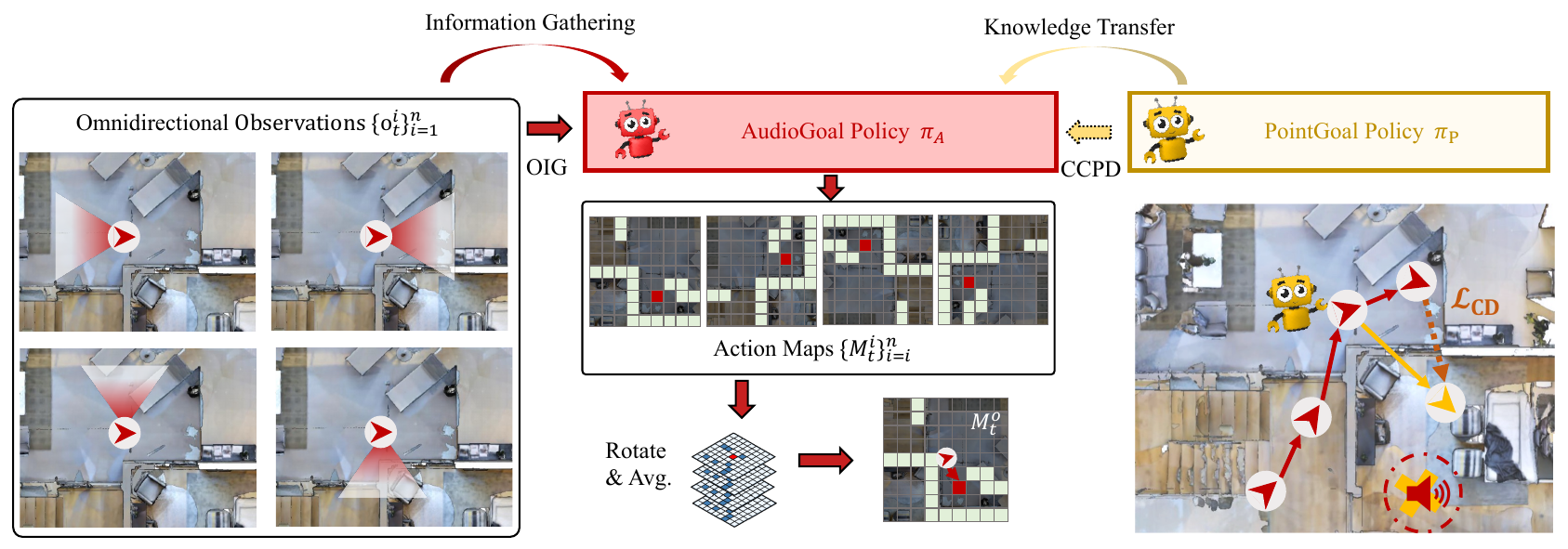}
   \caption{The overview of \texttt{ORAN}. There are two main components of the proposed \texttt{ORAN}, \ie OIG (in left) and CCPD (in right). The CCPD transfers the navigation knowledge from a PointGoal agent during training, and OIG gathers omnidirectional information during inference. The OIG and CCPD together make the \texttt{ORAN} a powerful AudioGoal agent.}\label{fig:pipline}
\end{center}

\end{figure*}

\noindent\textbf{Problem Setup.} 
In the AudioGoal navigation task~\cite{soundspaces, AudioVisualNavigation00}, the agent is initially located at position $p_0$ in a 3D environment that is partially observable. The agent's objective is to navigate to a persistent sounding target located at $p_\tau$. This task is typically formulated as a partially observable Markov decision process (POMDP) represented by a 7-tuple, $\{ \mathcal{S}, \mathcal{A},\Omega, O, \mathcal{T}, \mathcal{R}, \gamma \}$, where $\mathcal{S}$, $\mathcal{A}$, and $\Omega$ denote sets of states, actions, and observations, respectively.
The observation function $o_t=O(s_{t})$ maps the current state to an observation, where $o_t\in\Omega$ and $s_t\in \mathcal{S}$.
The state transition function $s_{t+1}= \mathcal{T}(s_t, a_t)$ describes the evolution of the system, where $a_t \in \mathcal{A}$. The reward function $r_t = \mathcal{R}(s_t,a_t)$ assigns a scalar reward to the agent's action at time $t$, and $\gamma$ is the discount factor used to weight future rewards. The navigation policy $\pi : \Omega \rightarrow \triangle^{\vert \mathcal{A} \vert}$ outputs the distribution of actions based on the current observation.
The training target is to approach the optimal policy $\pi^*$:
\begin{equation}
\small
    \pi^*=argmax_\pi \lbrack \mathbb{E}_\pi ( \Sigma_{t=1}^{\vert \tau \vert} \gamma^{t-1}r_t ) \rbrack.
\end{equation}
\noindent\textbf{Basic Architecture.} 
As depicted in Figure~\ref{fig:network}, the basic process of navigation can be formulated as the iterative execution of the following three steps~\cite{setwaypoints}:
\vspace{1mm}
\begin{itemize}[leftmargin=*]
	\setlength{\itemsep}{0pt}
	\setlength{\parsep}{-2pt}
	\setlength{\parskip}{-0pt}
	\setlength{\leftmargin}{-10pt}
	\vspace{-5pt}
	\item \textit{Observation $O(s_t)$}: The agent constructs top-down geometric and acoustic maps of the observed environment using egocentric visual-audio perceptions. At step $t$, the depth observation is back-projected to the 3D point cloud and then transformed into a 2D local map, which is used to update a \textit{global geometry map} $g_t$. $g_t$ has two channels, one for recording explored/unexplored areas and the other for occupied/free areas. Similarly, based on binaural audio perception $b_t$, a \textit{global acoustic map} $i_t$ is updated, which stores all sound intensities and their corresponding locations. The observation function can be formulated as:
     \begin{equation}
        \small
                 o_t = O(s_t) = (g_t, i_t, b_t).
    \end{equation}

    \item \textit{Decision $\pi_A(o_t)$}: A parameterized AudioGoal policy $\pi_{A}$ predicts the next short-term navigation goal $a_t$ based on multimodal cues and partial maps. We apply $\pi_{A}$ as a GRU-based waypoint predictor to approximate $\pi^*$, which generates the next short-term navigation goal. $\pi_{A}$ takes the current observation $o_t$ as input and outputs a score map $M_t$ of size $m \times m$, representing the probability distribution for the next waypoint:
    \begin{equation}
    \small
    M_t = \pi_{A}(o_t).
    \end{equation}
$M_t$ represents a set of $m \times m$ candidate waypoints and their associated action probabilities within a region centered around the agent. The waypoint predictor, denoted as $\pi_{A}$, is trained using reinforcement learning to predict actions at the waypoint level. Essentially, $\pi_{A}$ serves as the AudioGoal navigation policy for action prediction. we sample the next waypoint $a_t$ from $M_t$ based on the probability predicted by $\pi_A$.

\vspace{1mm}
	\item \textit{State Transition $\mathcal{T}(s_t,a_t)$}: The agent plans the shortest feasible path to the waypoint and executes the plan using low-level actions to obtain the new navigation state $s_{t+1}$. After selecting the waypoint $a_t$, the agent computes the shortest path from its current position to $a_t$ using Dijkstra's algorithm. It then moves towards $a_t$ by executing a sequence of low-level actions. 
 This process is repeated until the agent either selects its standing location as the next waypoint or reaches a navigation step limit.
	\vspace{-3pt}
\end{itemize}

Based on this waypoint-based scheme, \texttt{ORAN} further improves its waypoint navigation policy through CCPD (\S\ref{sec:CCPD}), and changes its forward-only perception with 90-degree field of view to an omnidirectional visual-audio information gathering mode -- OIG  (\S\ref{sec:OIG}).

\subsection{Confidence-Aware$_{\!}$ Cross-task$_{\!}$ Policy$_{\!}$ Distillation$_{\!\!\!}$}\label{sec:CCPD}

The PointGoal navigation policy is a valuable source of knowledge for indoor navigation. To transfer this knowledge, we propose using CCPD, which leverages a PointGoal policy ($\pi_{P}$) as a teacher to guide the training process of an AudioGoal policy ($\pi_{A}$). In each episode, the target $p_{\tau}$ for $\pi_P$ and $\pi_A$ is set to the same position, allowing them to share the same optimal policy $\pi^*$.
Since $\pi_{P}$ is trained using a large-scale training samples and achieves near-perfect performance, it provides a more accurate estimation of $\pi^*$.
So we train $\pi_{A}$ to imitate the actions of $\pi_{P}$, using policy distillation. The training target can be defined as follows:
\begin{equation}
\small
\pi_{A}^{*} = argmin_{\pi_A}\mathbb{E}_\tau [\mathcal{D} ( \pi_{A}, \pi_{P} )],
\end{equation}
where $\mathcal{D}(\cdot,\cdot)$ is a distance metric function, and $\tau=\{o_1,o_2,\dots,o_{|\tau|}\}$ are the observations from a sampled trajectory via a control policy. We utilize $\pi_A$ as the control policy to implement student-forcing sampling, which helps to reduce the distribution distance between training and inference, as mentioned in multiple works~\cite{czarnecki2019distilling,DueDis}. We set $\mathcal{D}(\cdot,\cdot)$ as the KL divergence function and the loss function of policy distillation can be defined as:
\begin{equation}
\small
 \mathcal{L}_D= \mathbb{E}_\tau \lbrack \mathcal{D}_{KL} (\pi_{A}, \pi_{P}) \rbrack.\label{eq:base-PD}
\end{equation}
However, the intrinsic differences between the AudioGoal and PointGoal tasks create a domain gap between the policies $\pi_A$ and $\pi_P$. Applying distillation supervision $\mathcal{L}_D$ over all actions in a trajectory can introduce unnecessary bias due to this gap. To mitigate this side effect, we propose selecting a subset of steps for distillation based on the action confidence of $\pi_P$, \ie the confidence-aware reweighting mechanism.
We measure the confidence of $\pi_{P}$ using the Shannon entropy, which is defined as:
\begin{equation}
\small
\mathcal{H}[\pi_{P}(o_t)] = -\sum \pi_{P}(o_t)\log \pi_{P}(o_t).
\end{equation}
Intuitively, the entropy of the action distribution increases as it becomes more even, indicating that the agent is less confident in choosing a particular action. Therefore, we choose the steps with lower entropy (\ie, higher confidence) for distillation. We calculate the confidence factor of $\pi_P$ as:
\begin{equation}
\small
    c_P(o_t) = \frac{1}{\mathcal{H}[\pi_{P} (o_t)]},
\end{equation}
The actions with higher $c_P(o_t)$ are crucial for navigation and less risk for $\pi_{A}$ to follow. 
We rank the $o_t$ in $\tau$ based on $c_P(o_t)$, and take the top $k$ to compute the distillation loss. So the loss with the confidence-aware reweighting can be defined as:
\begin{equation}
\small
 \mathcal{L}_{CD}= \mathbb{E}_\tau \{ \sum \mathbb{I}_k[c_P(o_t)]c_P(o_t)\mathcal{D}_{KL}[\pi_{A}(o_t), \pi_{P}(o_t)]\}.
 \label{eq:CCPD}
\end{equation} 
The notation $\mathbb{I}_k(\cdot)$ denotes selecting the top $k$ largest elements. In the training phase, we employ a combination of CCPD and PPO~\cite{ppo} to train the agent. 
With CCPD, the $\pi_{A}$ can learn the wayfinding ability from $\pi_{P}$ efficiently and performs better with less training samples, as shown in \S\ref{sec:experiments}.

\subsection{Omnidirectional Information Gathering}\label{sec:OIG}
Another notable benefit of \texttt{ORAN} is its ability to facilitate an omnidirectional visual-acoustic information gathering process. 
As mentioned in \S\ref{problem+network}, the decision model $\pi_A$ presently relies solely on visual-audio information obtained from a single direction to set the subsequent intermediate goals through the self-centered action distribution map $M_t$.
The $\mathcal{T}(s_t,a_t)$, which includes a non-parametric path planner, enables the agent to progress towards the next waypoint. Owing to the randomness in $\mathcal{T}(s_t,a_t)$, the 
agent's orientation on the subsequent waypoint is rendered indeterminate. So changing the agent's direction has limited impact on $\pi_A$ while predicting the next waypoint, and the decision-making mode of $\pi_A$ is omnidirectional.
However, the forward-only perception contradicts the decision-making mode, making the intermediate goal unreliable for two reasons. 
Firstly, the agent's perception is limited in range, resulting in incomplete occupancy maps and audio information. This limitation impedes the agent's ability to understand its surrounding environment, thereby reducing the navigation's robustness. Secondly, in our experiments, we have observed that the predicted $M_t$ exhibits a direction-relative bias. For instance, when the agent is uncertain about the direction to take, it may tend to gravitate towards a particular waypoint on $M_t$, potentially resulting in the agent becoming trapped in certain positions.
To overcome these obstacles, we enable the agent to collect information and integrate action decisions from multiple directions, as depicted in Figure \ref{fig:pipline}.

At each action step of $\pi_{A}$, we collect $n$ observations $\{o_t^{i}\}_{i=1}^n $ from $n$ directions in $\mathbf{\Omega}=\{ \omega_i\}_{i=1}^n$ to obtain panoramic information. Next we input them to $\pi_{A}$ to obtain the action maps $\{M^i_t\}_{i=1}^n$ in all directions :
\begin{equation}
\small
M^i_t = \pi_{A}(o_t^{i}).
\end{equation}
In order to collect the action distribution of various orientations, we rotate the action map to a consistent direction and then aggregate the distribution. Therefore, the resulting omnidirectional action map $M_t^o$ can be expressed as follows:
\begin{equation}
    \small
    M^o_t =\frac{1}{n}\sum_{i \leq n } R(M^i_t,-\omega_i),
\end{equation}
where $R(X,\theta)$ represents the rotation of the $X$ matrix for $\theta$ degree. the next waypoint is sampled base on the distribution of $M^o_t$.
To enhance the agent's ability to make accurate terminal judgments, we additionally employ a stop policy model to assist $\pi_A$ in identifying the target position.
With this OIG, the \texttt{ORAN} agent exhibits greater accuracy and efficiency in navigation without requiring finetuning. 
\begin{table*}[htb]
   \begin{center}
      \resizebox{\linewidth}{!}{
\input{tables/sota}
      }
   \end{center}
      \vspace{-4pt}
   \caption{\textbf{Comparisons on Soundspaces dataset}. Comparison with the state-of-the-art methods on the Soundspaces challenge dataset. The proposed \texttt{ORAN} boosts the performance in terms of all the key metrics on the unheard setting.
   }
      \vspace{-6pt}
    \label{tab:sota}
\end{table*}

\section{Experiments}\label{sec:experiments}



\subsection{Experimental Setup}
\vspace{1mm}
\noindent\textbf{Environments.}
We utilize the Soundspaces to evaluate our approach which encompasses two separate environments, namely Matterport3D~\cite{chang2017matterport3d} and Replica~\cite{replica}. To be consistent with prior work~\cite{soundspaces,setwaypoints}, we divide the scenes into three sets, train/val/test, consisting of $9/4/5$ for the Replica and $73/11/18$ for the Matterport3D. We evaluate the performance of our agent in both environments under two different settings following ~\cite{setwaypoints}: heard and unheard. 
The unheard setting in Matterport3D is the most demanding, and it is also our primary focus.
\begin{table*}[t]
      \resizebox{\linewidth}{!}{
         \input{tables/ablation_CCPD}
      }
      \vspace{2mm}
   \caption{\textbf{Ablations.} The ablation study on the key component of \texttt{ORAN}, see \S\ref{sec:ablation} for details.}
   \label{tab:ablations1}
   \vspace{-8pt}
\end{table*}

\noindent\textbf{Evaluation Metrics.}
We measure navigation performance using the following metrics:
1) Success Rate (SR): the ratio of agent trajectories where the agent stops at the goal position;
2) Success weighted by Path Length (SPL): Evaluating navigation efficiency involves factoring in the length of the agent's path and weighing its success accordingly;
3) SoftSPL: a version of SPL where binary success is replaced by progress towards the goal;
4) Success weighted by the Number of Actions (SNA): measures action efficiency via weighting success by the number of actions taken.

\noindent\textbf{Implementation Details.}
We set up the simulator following \cite{setwaypoints}. The PointGoal navigation model is initialized with the pre-trained model in \cite{wijmans2019dd}. We substitute the action prediction layer and finetune the model on the training split to predict the action map as that of $\pi_A$.
The learning rate is $2.5\times10^{-4}$ and decreases linearly during training. We set the hyper-parameter of the PPO~\cite{ppo} following~\cite{setwaypoints}.
The loss weight for $\mathcal{L}_{CD}$ is $0.3$.
We select the top $k=30$ most confident steps at each CCPD update step.
Both our method and baseline employ the spectrogram data augmentation during the training phase of the unheard split. In OIG, we specify the $\mathbf{\Omega}=\{0^{\circ}, 90^{\circ}, 180^{\circ}, 270^{\circ}\}$. The actions for information collection of OIG are not considered in metric as they are irrelevant to the decision-making process.

\vspace{1mm}
\noindent\textbf{Reproducibility.} Our model is implemented in PyTorch, and trained on one NVIDIA TITAIN RTX GPU.

\subsection{Comparison with State-of-the-art Methods}

 We compare our approach with existing methods, and the specifics of these methods are elaborated in the supplementary materials. Note that the metric used for ranking is the SPL on the unheard setting of Matterport3D in the Soundspaces 2022 challenge~\cite{soundspaces}. 
As shown in Table~\ref{tab:sota}, our model attains comparable performance to the AV-WaN model~\cite{setwaypoints} on the heard setting. Given the high SR observed in both environments on this setting, we primarily concentrate on comparing the performance on the unheard setting. In the Matterport3D environment, The \texttt{ORAN} model consistently outperforms all existing methods across all evaluation metrics. Specifically, \texttt{ORAN} achieves a SR of $59.4\%$, surpassing the second-best AV-WaN~\cite{setwaypoints} approach by $2.7\%$. Additionally, \texttt{ORAN} demonstrates significant improvements in navigation efficiency, with a performance increase of $9.9\%$ in SPL. In the Replica environment, \texttt{ORAN} outperforms previous methods by a wide margin, achieving results of ($46.7\%$, $60.9\%$, $36.5\%$) in (SPL, SR, SNA) respectively, with a relative improvement of ($35\%$, $15\%$, $35\%$) over AV-WaN~\cite{setwaypoints}. These experimental results effectively demonstrate the efficacy of the \texttt{ORAN} model.
\subsection{Ablation Study}\label{sec:ablation}
In this section, a set of ablation studies are conducted in the matterport3D unheard setting to verify the proposed components, as shown in Table~\ref{tab:ablations1}. Moreover, the design of the CCPD is also discussed in Table~\ref{tab:methods} and Table~\ref{tab:ablations2}.

\noindent\textbf{Ablations on CCPD}. We first examine the influence of the CCPD. As shown in Table~\ref{tab:ablation:reweighting}, comparing with the base agent, \#2 with the CCPD lifts SR and SPL from ($53.1\%$, $38.6\%$) to ($59.5\%$,  $45.1\%$) on the unheard setting of Matterport3D. 
Additionally, as shown in Table~\ref{tab:methods}, we train our agent to follow the oracle shortest path planner by randomly selecting $30$ steps and computing the cross-entropy loss. The agent performs $3\%$ lower in SR and $2\%$ lower in SPL than trained with CCPD. This indicates that the knowledge transmitted via CCPD enhances fundamental navigational proficiency.
Moreover, we also attempt to directly apply the PointGoal agent for the AudioGoal task by introducing an audio-based direction predictor to provide pseudo-GPS for the PointGoal agent.
From Table~\ref{tab:methods}, we can observe that PointGoal + pseudo-GPS performs poorly in all metrics, which experimentally indicates the domain gap between PointGoal and AudioGoal, as another confirmation of the necessity of CCPD.
\begin{table}[t]
   \begin{center}\resizebox{\linewidth}{!}{
   \input{tables/methods.tex}
   }
      \caption{\textbf{Other Analysis of CCPD}. The comparison of agents trained with CCPD, shortest path oracle supervision, and PointGoal agent navigating with pseudo-GPS predictor in the Matterport3D unheard setting.} \label{tab:methods}
   \end{center}
   \vspace{-6pt}
\end{table}

\begin{table}
   \begin{center}
   \input{tables/ablation2}
   \end{center}
   \vspace{-3mm}
   \caption{\textbf{Influence of Selected Steps.} The models are evaluated in the Matterport3D unheard setting. The model performs best with top 30 most confident steps.}
   \label{tab:ablations2}  \vspace{-4pt}
\end{table}

\vspace{1mm}
\noindent\textbf{Ablations on OIG}. 
Our proposed \texttt{ORAN} model incorporates the OIG technique, which allows the agent to utilize information from different directions, leading to a more robust navigation performance. As shown in Table~\ref{tab:ablation:rate} \#2, the model with OIG achieves a significant improvement in both SPL and SoftSPL, from ($45.1\%$, $49.0\%$) to ($50.8\%$, $55.7\%$). This finding suggests that OIG enhances navigation efficiency in terms of distance traveled.
We follow previous efforts~\cite{setwaypoints} to view SPL as the primary ranking metric. It is worth mentioning that SPL and most metrics (\eg NE, SR, and SoftSPL) are unaffected by the perception actions for OIG. The OIG also improves the PointGoal policy in Table~\ref{tab:methods}, which achieves an SR of 95.1\% and an SPL of 74.3\%.

\vspace{1mm}
\noindent\textbf{Confidence-Aware Reweighting}. To confirm the essentiality of confidence-aware reweighting, we implement the basic policy distillation loss $\mathcal{L}_D$ described in Equation~\ref{eq:base-PD} in Table~\ref{tab:ablation:reweighting} \#1.
Comparing \#2 to \#1, the model trained with $\mathcal{L}_{CD}$ improves SR and SPL from ($50.0\%$, $37.0\%$) to ($59.5\%$, $45.1\%$). We can observe that the SR and SPL metrics show a decrease when trained without the reweighting, as compared to the baseline model. This outcome exemplifies the significance of employing confidence-based selection and reweighting during training.
\begin{figure}[t]
   \begin{center}
      \includegraphics[width=1\linewidth]{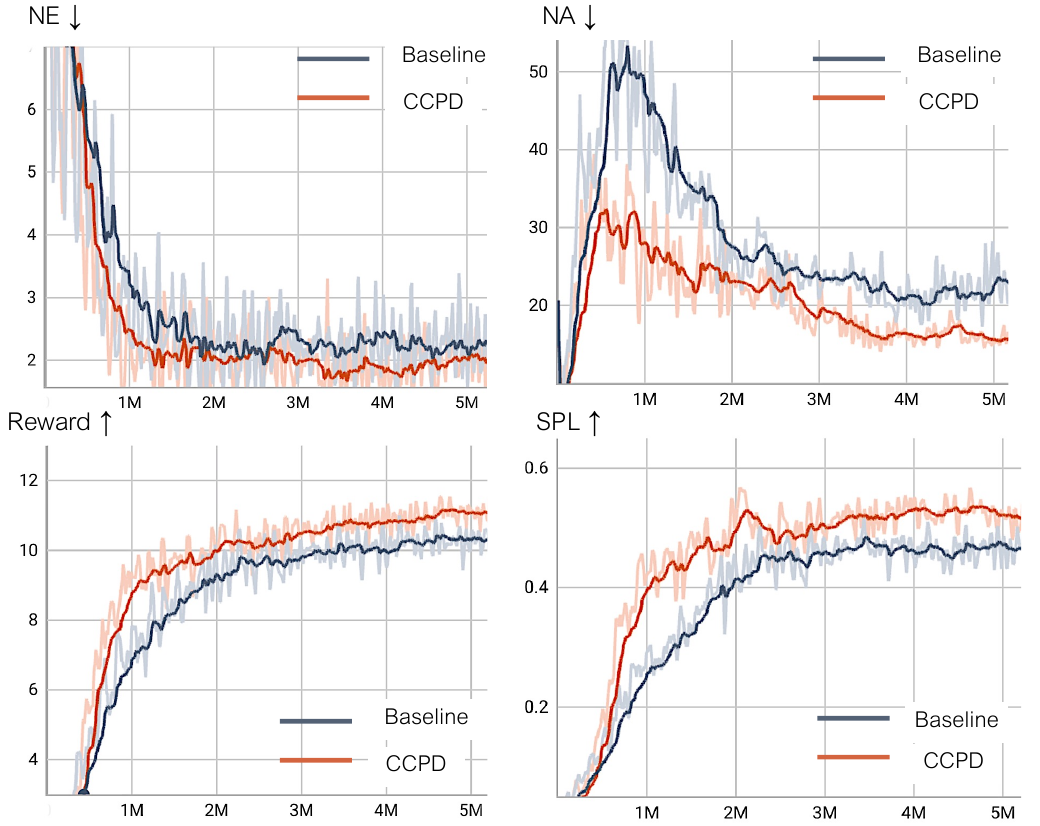}
   \end{center}
   \caption{\textbf{Qualitative Analysis for CCPD.} The navigation metrics vary with the training steps. The model with CCPD exhibits faster convergence and superior performance.}\label{fig:vis1}
\end{figure}

\vspace{1mm}
\noindent\textbf{Selected steps $k$ in CCPD}. 
In Table~\ref{tab:ablations2}, we investigate the impact of the number of selected steps in CCPD, denoted by $k$, on the performance of the model. Specifically, we compare the performance of the model trained with the top $k=$ $10$, $30$, $50$, and $100$ most confident steps. Notably, we observe that the model attains the best performance at $k=30$, whereas the navigation performance decreases with an increase in $k$ to 50 or 100. This implies that supplementary supervision does not yield any positive effects.
\begin{figure*}[t]
   \begin{center}
      \includegraphics[width=0.9\linewidth]{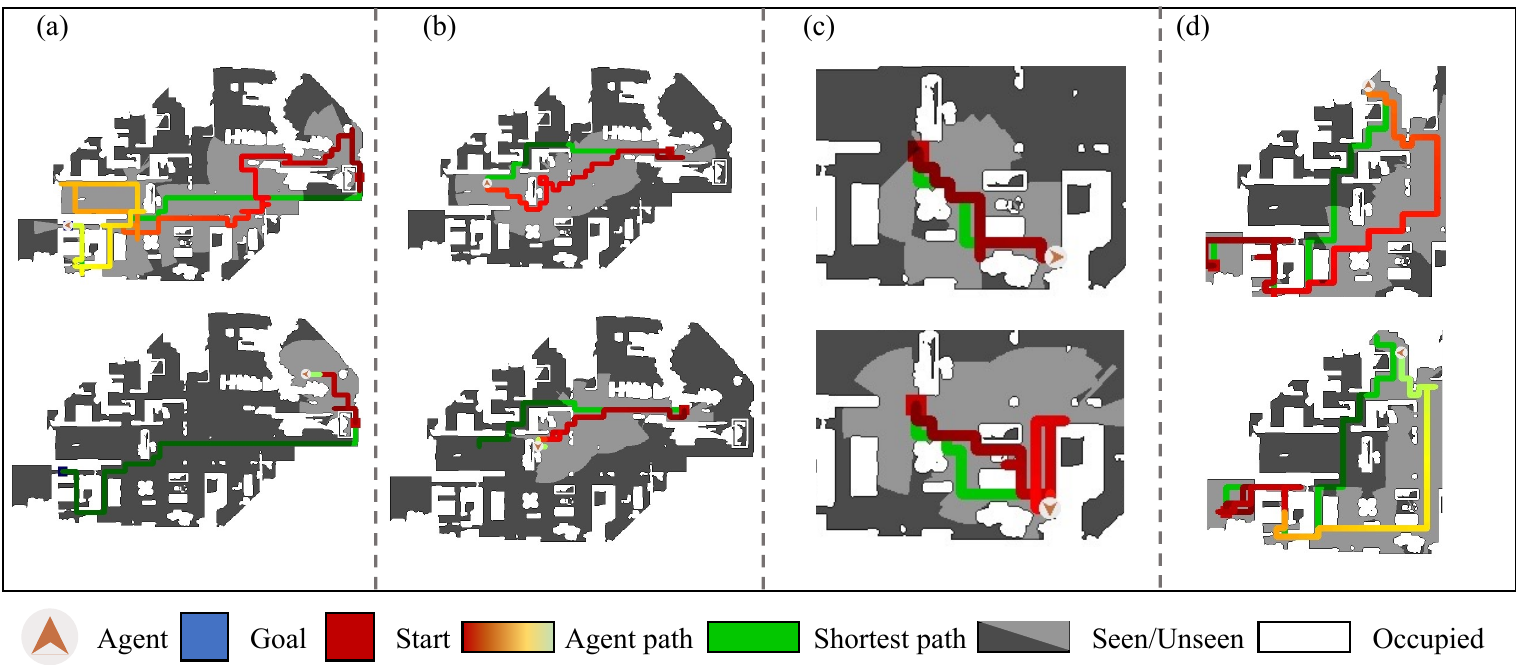}
   \end{center}
   \caption{\textbf{Qualitative Analysis for CCPD.} The navigation trajectories of the agent trained with CCPD (in the first row) or not (in the second row). Each column is of the same test episode. The agent with CCPD shows more robust navigation performance, especially on long trajectories. The color of the agent's path turns from red to green gradually to indicate the trajectory length.}\label{fig:vis-ccpdt}
\label{fig:CCPD}
\end{figure*}

\begin{figure*}[t]
   \begin{center}
      \includegraphics[width=0.9\linewidth]{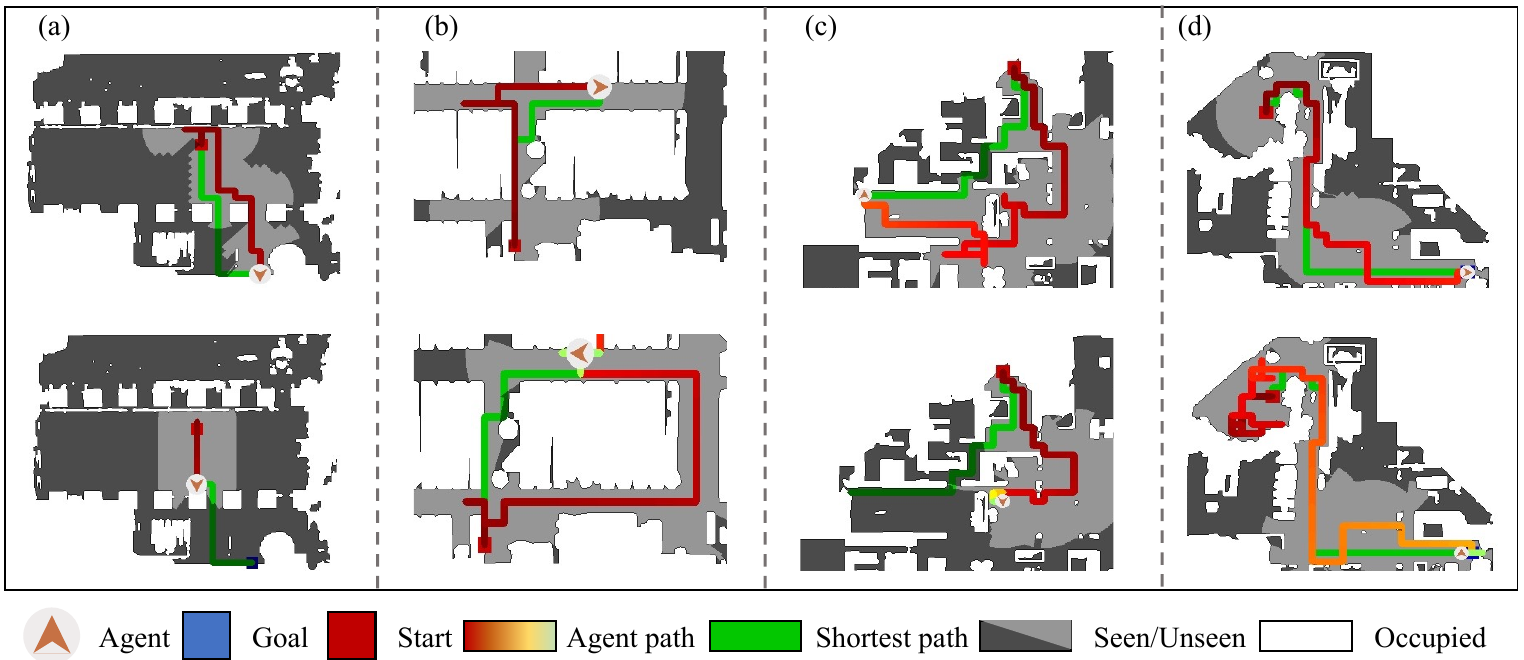}
   \end{center}
   \caption{\textbf{Qualitative Analysis for OIG.} The navigation trajectories of the same agent with OIG (in the first row) or not (in the second row). Each column corresponds to the same test episode. The gradual transition of the path color from red to green denotes the trajectory length.
}\label{fig:vis2}
\end{figure*}

\begin{table*}[t]
      \begin{center}
          \resizebox{\linewidth}{!}{
         \input{tables/ensemble.tex}

         \vspace{-3mm}
      }
      \end{center}      
   \caption{\textbf{Model ensemble}. Fusing model with different accuracy can increase the navigation accuracy. OIG also works on the Fusion model.} \label{tab:ensemble}
   \vspace{-1mm}
\end{table*}
\subsection{Qualitative Analysis}\label{sec:vis}
\noindent\textbf{Qualitative Analysis for CCPD}.
 In Figure~\ref{fig:vis1}, 
 we investigate the influence of CCPD on the training phase of the AudioGoal agent by visualizing the curves of four navigation metrics — namely, NE (Navigation Error), NA (Navigation Actions), reward, and SPL — as they vary with the training steps.
The model trained with CCPD achieves a lower level of NA and NE faster than the baseline model. During the initial training stage ($0$ $\rightarrow$ $1$M steps), the reward and SPL of the model trained with CCPD grow faster and maintain a higher level throughout the remainder of the training phase.
This demonstrates that CCPD, as a powerful knowledge transfer algorithm, helps the agent to better master AudioGoal navigation with far fewer training samples.
Figure~\ref{fig:vis-ccpdt} depicts the comparison of the model trained with CCPD (in the first row) or not (in the second row) of the same episodes. We can observe that the model trained with CCPD performs better navigation robustness and efficiency, especially on the episodes with longer trajectories.
\vspace{1mm}

\noindent\textbf{Qualitative Analysis for OIG}. To analyze how the omnidirectional information benefits the navigation, we visualize the same agent's trajectories equipped with OIG or not. We list here some key observations drawn from Figure~\ref{fig:vis2}:
First, the agent with OIG can distinguish whether arrive at the sounding target more precisely, as in episode (a) and avoid hanging around the target, as in episode (b). 
Second, OIG gets the agent out of the deadlock, as in episode (c). 
Third, the agent with OIG is more likely to find the right direction towards the target and reduce detours, as in episodes (b) and (d). 
These phenomena intuitively show the help of omnidirectional information to the agent's navigation.

\subsection{Model Ensemble}\label{sec:ensemble}
We apply post-fusion on models with diverse network architectures and training strategies to push the limits of the AudioGoal task.
At each time step, we feed the current observation into each of these models and then average the resulting action maps from all models to obtain the final action distribution.
Using a grid search strategy, we select a combination of trained models that yield promising results. 
 As shown in Table~\ref{tab:ensemble}, the selected models achieved SPL percentages of $40.9\%$, $43.5\%$, $42.4\%$, and $43.1\%$, respectively. Further details about these models are provided in the supplemental material. Notably, we observe that as we increase the number of models from \#1 to \#4, the gain in performance obtained from the model ensemble decreases.
By employing the post-fusion strategy, our agent achieves an improvement of 6.5\% in SPL accuracy over the best single model (from $43.5\%$ to $50.0\%$). Additionally, 
compared to model \#4, model \#5 incorporating OIG technology yields significant enhancements in accuracy, with absolute improvements of ($12.6\%$, $11.6\%$, $5.0\%$, and $12.5\%$) observed in SPL, SR, SNA, and SoftSPL metrics, respectively. These results highlight the inadequacy of forward-only perception in providing crucial information required for the AudioGoal wayfinding process.

\section{Conclusion}
We present an omnidirectional audio-visual navigator based on cross-task navigation skill transfer, named \texttt{ORAN}, that advances state-of-the-art technologies for AudioGoal in two aspects. Firstly, we transfer the navigation knowledge from the PointGoal policy via a cross-task policy distillation to sharpen the basic point-to-point wayfinding ability. 
We further consider the action confidence of the PointGoal agent to guide the distillation to overcome the domain gap between the two tasks.
Secondly,  we equip the AudioGoal agent with the ability to gather information from different directions, instead of the only direction it is facing, to better support the omnidirectional decision-making.
Moreover, after model assembling, \texttt{ORAN} yields $12\%$ absolute promotion in SPL, which makes the champion solution of the Soundspaces challenge in 2022. 

\noindent\textbf{Acknowledgements.} This work is supported in part by the National Key R\&D Program of China under Grant 2022ZD0115502, the National Natural Science Foundation of China under Grant 62122010, the Fundamental Research Funds for the Central Universities, the Fundamental Research Funds for the Central Universities (No. 226-2022-00051), and CCF-Tencent Open Fund. 

{\small
\bibliographystyle{ieee_fullname}
\bibliography{egbib}
}

\end{document}

%% file: tables/sota.tex
\setlength\tabcolsep{8pt}
\begin{tabular}{r|ccc|ccc|ccc|ccc} 
\toprule
                         & \multicolumn{6}{c|}{\textit{Replica}}                                                                                                               & \multicolumn{6}{c}{\textit{Matterport3D}}                                                                                                                              \\ 
\cline{2-13}
                         & \multicolumn{3}{c|}{\textit{Heard}}           & \multicolumn{3}{c|}{\textit{Unheard}}                                                               & \multicolumn{3}{c|}{\textit{Heard}}                             & \multicolumn{3}{c}{\textit{Unheard}}                                                                 \\
Model                    & SNA           & SR            & SPL           & SNA                    & SR            & {\cellcolor[rgb]{0.941,0.941,0.941}}SPL                    & SNA                    & SR            & SPL                    & SNA                    & SR            & {\cellcolor[rgb]{0.941,0.941,0.941}}SPL                     \\ 
\shline
Random Agent~\cite{setwaypoints}        & 1.8           & 18.5          & 4.9           & 1.8                    & 18.5          & {\cellcolor[rgb]{0.941,0.941,0.941}}4.9                    & 0.8                    & 9.1           & 2.1                    & 0.8                    & 9.1           & {\cellcolor[rgb]{0.941,0.941,0.941}}2.1                     \\
Direction Follower~\cite{setwaypoints}      & 41.1          & 72.0          & 54.7          & 8.4                    & 17.2          & {\cellcolor[rgb]{0.941,0.941,0.941}}11.1                   & 23.8                   & 41.2          & 32.3                   & 10.7                   & 18.0          & {\cellcolor[rgb]{0.941,0.941,0.941}}13.9                    \\
Frontier Waypoints ~\cite{setwaypoints}     & 35.2          & 63.9          & 44.0          & 5.1                    & 14.8          & {\cellcolor[rgb]{0.941,0.941,0.941}}6.5                    & 22.2                   & 42.8          & 30.6                   & 8.1                    & 16.4          & {\cellcolor[rgb]{0.941,0.941,0.941}}10.9                    \\
Supervised Waypoints~\cite{setwaypoints}   & 48.5          & 88.1          & 59.1          & 10.1                   & 43.1          & {\cellcolor[rgb]{0.941,0.941,0.941}}14.1                   & 16.2                   & 36.2          & 21.0                   & 2.9                    & 8.8           & {\cellcolor[rgb]{0.941,0.941,0.941}}4.1                     \\
Gan et al. ~\cite{setwaypoints}            & 47.9          & 83.1          & 57.6          & 5.7                    & 15.7          & {\cellcolor[rgb]{0.941,0.941,0.941}}7.5                    & 17.1                   & 37.9          & 22.8                   & 3.6                    & 10.2          & {\cellcolor[rgb]{0.941,0.941,0.941}}5.0                     \\
AV-Nav~\cite{soundspaces}                  & 52.7          & 94.5          & 78.2          & 16.7                   & 50.9          & {\cellcolor[rgb]{0.941,0.941,0.941}}34.7                   & 32.6                   & 71.3          & 55.1                   & 12.8                   & 40.1          & {\cellcolor[rgb]{0.941,0.941,0.941}}25.9                    \\
AV-WaN~\cite{setwaypoints}                  & 70.7 & 98.7 & 86.6 & 27.1                   & 52.8          & {\cellcolor[rgb]{0.941,0.941,0.941}}34.7                   & 54.8                   & 93.6 & 72.3                   & 30.6                   & 56.7          & {\cellcolor[rgb]{0.941,0.941,0.941}}40.9                    \\ 
\hline
\texttt{ORAN} (\textbf{ours})     & 70.1          & 96.7          & 84.2          & \textbf{\textbf{36.5}} & \textbf{60.9} & {\cellcolor[rgb]{0.941,0.941,0.941}}\textbf{\textbf{46.7}} & 57.7 & 93.5          & 73.7 & \textbf{\textbf{35.3}} & \textbf{59.4} & {\cellcolor[rgb]{0.941,0.941,0.941}}\textbf{\textbf{50.8}}  \\
\bottomrule
\end{tabular}

%% file: tables/ablation_CCPD.tex
\subfloat[ \textbf{Detail analysis on essential components of \texttt{ORAN}}, \ie CCPD and OIG on the Matterport3D unheard setting. The performance is gradually improved with the continuous addition of proposed modules. \label{tab:ablation:rate}]{
\begin{tabular}{rr|c|c|cccccc}
  \toprule
  & \multirow{1}{*}{Name} &   \multirow{1}{*}{CCPD} &\multirow{1}{*}{OIG}   & 
  SR$\uparrow$         & SPL$\uparrow$           & SoftSPL$\uparrow$     \\
  \hline
  &  Baseline       &                             &                            & 53.1    &   38.6    &  45.0  \\    
    &   \#1                  & \checkmark                  &                    & \textbf{59.5}    &  45.1    &  49.0  \\     
    &   \#2                  & \checkmark                  &  \checkmark        & 59.4    &  \textbf{50.8}    &  \textbf{55.7}   \\   
  \bottomrule
\end{tabular}
}\hspace{4mm}
\subfloat[\textbf{Ablation analysis on CCPD}. The ablation study on the effect of confidence-aware reweighting on the Matterport3D unheard setting. The Confidence-aware reweighting gains a large accuracy improvement.    \label{tab:ablation:reweighting}]{
\begin{tabular}{rr|c|c|ccc}
  \toprule
  & \multirow{1}{*}{Name} &   \multirow{1}{*}{$\mathcal{L}_{D}$} &\multirow{1}{*}{$\mathcal{L}_{CD}$}   & 
  SR$\uparrow$         & SPL$\uparrow$           & SoftSPL$\uparrow$     \\
  \hline
  &  Baseline       &                             &                            & 53.1    &   38.6    &  45.0  \\    
    &   \#1                  & \checkmark                  &                    & 50.0    &  37.0    &  43.2  \\  
    &   \#2                  &                   & \checkmark                    & \textbf{59.5}    &  \textbf{45.1}    &  \textbf{64.0}  \\  

  \bottomrule
\end{tabular}
}

%% file: tables/methods.tex
\setlength\tabcolsep{8pt}
\begin{tabular}{r|ccccc}
\toprule

   Models & SR         & SPL           & SoftSPL    \\\shline
    CCPD  & \textbf{59.5}   &  \textbf{45.1}  &  \textbf{49.0}           \\  
    Oracle Supervision  &   56.7   &    42.2      &  45.7         \\ 
    PointGoal +  Pseudo-GPS & 17.2    &  12.3    & 12.1         \\ 
    \hline
    PointGoal & 94.4      &  74.2    & 74.4     \\  
\bottomrule
\end{tabular}

%% file: tables/ablation2.tex
\resizebox{1\linewidth}{!}{
\setlength\tabcolsep{19pt}
\begin{tabular}{c|ccc}
  \toprule

 $k$ & SR$\uparrow$         & SPL$\uparrow$           & SoftSPL$\uparrow$    \\\shline
     {10} & 57.7      & 37.1       & 41.8           \\
     {30} & \textbf{59.5}      & \textbf{45.1}      & \textbf{49.0}     \\ 
    {50} & 56.1       & 39.6       & 44.1         \\ 
    {100}  & 51.2       & 37.3       & 42.5       \\  
    \bottomrule
\end{tabular}
}

%% file: tables/ensemble.tex
\setlength\tabcolsep{12pt}
\begin{tabular}{c|c|c|c|c|c|cccc}
  \toprule
 \multirow{2}{*}{\#}  & \multirow{2}{*}{OIG}   & \multicolumn{4}{c|}{{\textit{Fused Models}}}    & \multicolumn{4}{c}{{\textit{Matterport3D Unheard}}} \\
\cline{3-10}
& &  \texttt{A:40.9}      & \texttt{B:43.5}      &  \texttt{C:42.4}     & \texttt{D:43.1}  & SPL$\uparrow$  & SR$\uparrow$ & SNA$\uparrow$ & SoftSPL$\uparrow$ \\
  \shline
1 & & \checkmark      &      &       &       &    40.9    &   56.7  &  30.6  & 46.3   \\
2 & & \checkmark      & \checkmark      &       &       &    47.5    &   66.1  &  35.3  & 51.7   \\
3 & & \checkmark      & \checkmark      &  \checkmark     &       &  48.0  &   64.7   & 35.9   & 53.7           \\
4& & \checkmark      & \checkmark      &  \checkmark     &  \checkmark &  50.0  & 66.1 & 37.2 & 54.7  \\
5&\checkmark   & \checkmark      & \checkmark      &  \checkmark     &  \checkmark       &   \textbf{62.6} & \textbf{77.7} & \textbf{42.2} & \textbf{67.2}  \\
    \bottomrule
\end{tabular}

%% file: ICCV.bbl
\begin{thebibliography}{10}\itemsep=-1pt

\bibitem{al2022zero}
Ziad Al-Halah, Santhosh~Kumar Ramakrishnan, and Kristen Grauman.
\newblock Zero experience required: Plug \& play modular transfer learning for semantic visual navigation.
\newblock In {\em CVPR}, 2022.

\bibitem{an2021neighbor}
Dong An, Yuankai Qi, Yan Huang, Qi Wu, Liang Wang, and Tieniu Tan.
\newblock Neighbor-view enhanced model for vision and language navigation.
\newblock In {\em ACM MM}, 2021.

\bibitem{an2022bevbert}
Dong An, Yuankai Qi, Yangguang Li, Yan Huang, Liang Wang, Tieniu Tan, and Jing Shao.
\newblock Bevbert: Topo-metric map pre-training for language-guided navigation.
\newblock {\em arXiv preprint arXiv:2212.04385}, 2022.

\bibitem{anderson2018vision}
Peter Anderson, Qi Wu, Damien Teney, Jake Bruce, Mark Johnson, Niko S{\"u}nderhauf, Ian Reid, Stephen Gould, and Anton van~den Hengel.
\newblock Vision-and-language navigation: Interpreting visually-grounded navigation instructions in real environments.
\newblock In {\em CVPR}, 2018.

\bibitem{brown2013consciousness}
Richard Brown.
\newblock {\em Consciousness inside and out: Phenomenology, neuroscience, and the nature of experience}.
\newblock Springer Science \& Business Media, 2013.

\bibitem{3-sound-attack}
Nicholas Carlini and David~A. Wagner.
\newblock Audio adversarial examples: Targeted attacks on speech-to-text.
\newblock In {\em IEEE Security and Privacy Workshops}, 2018.

\bibitem{chang2017matterport3d}
Angel Chang, Angela Dai, Thomas Funkhouser, Maciej Halber, Matthias Niessner, Manolis Savva, Shuran Song, Andy Zeng, and Yinda Zhang.
\newblock Matterport3d: Learning from rgb-d data in indoor environments.
\newblock {\em arXiv preprint arXiv:1709.06158}, 2017.

\bibitem{ObjectGoal02}
Devendra~Singh Chaplot, Dhiraj Gandhi, Abhinav Gupta, and Russ~R. Salakhutdinov.
\newblock Object goal navigation using goal-oriented semantic exploration.
\newblock In {\em NeurIPS}, 2020.

\bibitem{ObjectGoal03}
Devendra~Singh Chaplot, Ruslan Salakhutdinov, Abhinav Gupta, and Saurabh Gupta.
\newblock Neural topological {SLAM} for visual navigation.
\newblock In {\em CVPR}, 2020.

\bibitem{chen2021semantic}
Changan Chen, Ziad Al-Halah, and Kristen Grauman.
\newblock Semantic audio-visual navigation.
\newblock In {\em CVPR}, 2021.

\bibitem{savn}
Changan Chen, Ziad Al{-}Halah, and Kristen Grauman.
\newblock Semantic audio-visual navigation.
\newblock In {\em CVPR}, 2021.

\bibitem{soundspaces}
Changan Chen, Unnat Jain, Carl Schissler, Sebastia Vicenc~Amengual Gari, Ziad Al-Halah, Vamsi~Krishna Ithapu, Philip Robinson, and Kristen Grauman.
\newblock Soundspaces: Audio-visual navigation in 3d environments.
\newblock In {\em ECCV}, 2020.

\bibitem{setwaypoints}
Changan Chen, Sagnik Majumder, Al-Halah Ziad, Ruohan Gao, Santhosh Kumar~Ramakrishnan, and Kristen Grauman.
\newblock Learning to set waypoints for audio-visual navigation.
\newblock In {\em ICLR}, 2021.

\bibitem{chen22soundspaces2}
Changan Chen, Carl Schissler, Sanchit Garg, Philip Kobernik, Alexander Clegg, Paul Calamia, Dhruv Batra, Philip~W Robinson, and Kristen Grauman.
\newblock Soundspaces 2.0: A simulation platform for visual-acoustic learning.
\newblock In {\em NeurIPS 2022}, 2022.

\bibitem{chen2022reinforced}
Jinyu Chen, Chen Gao, Erli Meng, Qiong Zhang, and Si Liu.
\newblock Reinforced structured state-evolution for vision-language navigation.
\newblock In {\em CVPR}, 2022.

\bibitem{cho2019efficacy}
Jang~Hyun Cho and Bharath Hariharan.
\newblock On the efficacy of knowledge distillation.
\newblock In {\em CVPR}, 2019.

\bibitem{4-sound-attack}
Moustapha Ciss{\'{e}}, Yossi Adi, Natalia Neverova, and Joseph Keshet.
\newblock Houdini: Fooling deep structured visual and speech recognition models with adversarial examples.
\newblock In {\em NeurIPS}, 2017.

\bibitem{czarnecki2019distilling}
Wojciech~M Czarnecki, Razvan Pascanu, Simon Osindero, Siddhant Jayakumar, Grzegorz Swirszcz, and Max Jaderberg.
\newblock Distilling policy distillation.
\newblock In {\em AISTATS}, 2019.

\bibitem{AudioVisualNavigation00}
Victoria Dean, Shubham Tulsiani, and Abhinav Gupta.
\newblock See, hear, explore: Curiosity via audio-visual association.
\newblock In {\em NeurIPS}, 2020.

\bibitem{deng2022anomaly}
Hanqiu Deng and Xingyu Li.
\newblock Anomaly detection via reverse distillation from one-class embedding.
\newblock In {\em CVPR}, pages 9737--9746, 2022.

\bibitem{2-sound-attack}
Tianyu Du, Shouling Ji, Jinfeng Li, Qinchen Gu, Ting Wang, and Raheem Beyah.
\newblock Sirenattack: Generating adversarial audio for end-to-end acoustic systems.
\newblock In Hung{-}Min Sun, Shiuh{-}Pyng Shieh, Guofei Gu, and Giuseppe Ateniese, editors, {\em Asia Conference on Computer and Communications Security}, 2020.

\bibitem{Online2019Marc}
Marc Fischer, Matthew Mirman, Steven Stalder, and Martin~T. Vechev.
\newblock Online robustness training for deep reinforcement learning.
\newblock {\em CoRR}, abs/1911.00887, 2019.

\bibitem{fried2018speaker}
Daniel Fried, Ronghang Hu, Volkan Cirik, Anna Rohrbach, Jacob Andreas, Louis-Philippe Morency, Taylor Berg-Kirkpatrick, Kate Saenko, Dan Klein, and Trevor Darrell.
\newblock Speaker-follower models for vision-and-language navigation.
\newblock In {\em NeurIPS}, 2018.

\bibitem{AudioVisualNavigation01}
Chuang Gan, Yiwei Zhang, Jiajun Wu, Boqing Gong, and Joshua~B. Tenenbaum.
\newblock Look, listen, and act: Towards audio-visual embodied navigation.
\newblock In {\em ICRA}, 2020.

\bibitem{Gao_2021_CVPR}
Chen Gao, Jinyu Chen, Si Liu, Luting Wang, Qiong Zhang, and Qi Wu.
\newblock Room-and-object aware knowledge reasoning for remote embodied referring expression.
\newblock In {\em CVPR}, 2021.

\bibitem{georgakis2022uncertainty}
Georgios Georgakis, Bernadette Bucher, Anton Arapin, Karl Schmeckpeper, Nikolai Matni, and Kostas Daniilidis.
\newblock Uncertainty-driven planner for exploration and navigation.
\newblock In {\em ICRA}. IEEE, 2022.

\bibitem{Georgakis_2022_CVPR}
Georgios Georgakis, Karl Schmeckpeper, Karan Wanchoo, Soham Dan, Eleni Miltsakaki, Dan Roth, and Kostas Daniilidis.
\newblock Cross-modal map learning for vision and language navigation.
\newblock In {\em CVPR}, 2022.

\bibitem{gordon2019splitnet}
Daniel Gordon, Abhishek Kadian, Devi Parikh, Judy Hoffman, and Dhruv Batra.
\newblock Splitnet: Sim2sim and task2task transfer for embodied visual navigation.
\newblock In {\em ICCV}, 2019.

\bibitem{gupta2017cognitive}
Saurabh Gupta, James Davidson, Sergey Levine, Rahul Sukthankar, and Jitendra Malik.
\newblock Cognitive mapping and planning for visual navigation.
\newblock In {\em CVPR}, 2017.

\bibitem{hong2011recurrent}
Y Hong, Q Wu, Y Qi, C Rodriguez-Opazo, and S Gould.
\newblock A recurrent vision-and-language bert for navigation. arxiv 2021.
\newblock {\em arXiv preprint arXiv:2011.13922}, 2021.

\bibitem{5-sound-attack}
Dan Iter, Jade Huang, and Mike Jermann.
\newblock Generating adversarial examples for speech recognition.
\newblock {\em Stanford Technical Report}, 2017.

\bibitem{ai2thor}
Eric Kolve, Roozbeh Mottaghi, Winson Han, Eli VanderBilt, Luca Weihs, Alvaro Herrasti, Daniel Gordon, Yuke Zhu, Abhinav Gupta, and Ali Farhadi.
\newblock Ai2-thor: An interactive 3d environment for visual ai.
\newblock {\em arXiv}, 2017.

\bibitem{DueDis}
Kwei-Herng Lai, Daochen Zha, Yuening Li, and Xia Hu.
\newblock Dual policy distillation.
\newblock In {\em IJCAI}, 2020.

\bibitem{lyu2022improving}
Yunlian Lyu, Yimin Shi, and Xianggang Zhang.
\newblock Improving target-driven visual navigation with attention on 3d spatial relationships.
\newblock {\em Neural Processing Letters}, 54(5), 2022.

\bibitem{Partsey_2022_CVPR}
Ruslan Partsey, Erik Wijmans, Naoki Yokoyama, Oles Dobosevych, Dhruv Batra, and Oleksandr Maksymets.
\newblock Is mapping necessary for realistic pointgoal navigation?
\newblock In {\em CVPR}, 2022.

\bibitem{qi2020reverie}
Yuankai Qi, Qi Wu, Peter Anderson, Xin Wang, William~Yang Wang, Chunhua Shen, and Anton van~den Hengel.
\newblock Reverie: Remote embodied visual referring expression in real indoor environments.
\newblock In {\em CVPR}, 2020.

\bibitem{qiao2022hop}
Yanyuan Qiao, Yuankai Qi, Yicong Hong, Zheng Yu, Peng Wang, and Qi Wu.
\newblock Hop: history-and-order aware pre-training for vision-and-language navigation.
\newblock In {\em CVPR}, 2022.

\bibitem{qiao2023hop+}
Yanyuan Qiao, Yuankai Qi, Yicong Hong, Zheng Yu, Peng Wang, and Qi Wu.
\newblock Hop+: History-enhanced and order-aware pre-training for vision-and-language navigation.
\newblock {\em IEEE Transactions on Pattern Analysis and Machine Intelligence}, 2023.

\bibitem{qu2022importance}
Xinghua Qu, Yew~Soon Ong, Abhishek Gupta, Pengfei Wei, Zhu Sun, and Zejun Ma.
\newblock Importance prioritized policy distillation.
\newblock In {\em ACM KDD}, 2022.

\bibitem{ObjectGoal04}
Nikolay Savinov, Alexey Dosovitskiy, and Vladlen Koltun.
\newblock Semi-parametric topological memory for navigation.
\newblock In {\em ICLR}, 2018.

\bibitem{habitat2019}
Manolis Savva, Abhishek Kadian, Oleksandr Maksymets, Yili Zhao, Erik Wijmans, Bhavana Jain, Julian Straub, Jia Liu, Vladlen Koltun, Jitendra Malik, Devi Parikh, and Dhruv Batra.
\newblock Habitat: {A} {P}latform for {E}mbodied {AI} {R}esearch.
\newblock In {\em ICCV}, 2019.

\bibitem{ppo}
John Schulman, Filip Wolski, Prafulla Dhariwal, Alec Radford, and Oleg Klimov.
\newblock Proximal policy optimization algorithms.
\newblock {\em arXiv preprint arXiv:1707.06347}, 2017.

\bibitem{replica}
Julian Straub, Thomas Whelan, Lingni Ma, Yufan Chen, Erik Wijmans, Simon Green, Jakob~J Engel, Raul Mur-Artal, Carl Ren, Shobhit Verma, et~al.
\newblock The replica dataset: A digital replica of indoor spaces.
\newblock {\em arXiv preprint arXiv:1906.05797}, 2019.

\bibitem{1-sound-attack}
Vinod Subramanian, Arjun Pankajakshan, Emmanouil Benetos, Ning Xu, SKoT McDonald, and Mark~B. Sandler.
\newblock A study on the transferability of adversarial attacks in sound event classification.
\newblock In {\em IEEE ICASSP}, 2020.

\bibitem{habitat2021}
Andrew Szot, Alex Clegg, Eric Undersander, Erik Wijmans, Yili Zhao, John Turner, Noah Maestre, Mustafa Mukadam, Devendra Chaplot, Oleksandr Maksymets, Aaron Gokaslan, Vladimir Vondrus, Sameer Dharur, Franziska Meier, Wojciech Galuba, Angel Chang, Zsolt Kira, Vladlen Koltun, Jitendra Malik, Manolis Savva, and Dhruv Batra.
\newblock Habitat 2.0: Training home assistants to rearrange their habitat.
\newblock In {\em NeurIPS}, 2021.

\bibitem{wang2022towards}
Hanqing Wang, Wei Liang, Luc~V Gool, and Wenguan Wang.
\newblock Towards versatile embodied navigation.
\newblock {\em NeurIPS}, 2022.

\bibitem{wang2022counterfactual}
Hanqing Wang, Wei Liang, Jianbing Shen, Luc Van~Gool, and Wenguan Wang.
\newblock Counterfactual cycle-consistent learning for instruction following and generation in vision-language navigation.
\newblock In {\em CVPR}, 2022.

\bibitem{wang2021structured}
Hanqing Wang, Wenguan Wang, Wei Liang, Caiming Xiong, and Jianbing Shen.
\newblock Structured scene memory for vision-language navigation.
\newblock In {\em CVPR}, 2021.

\bibitem{wang2020active}
Hanqing Wang, Wenguan Wang, Tianmin Shu, Wei Liang, and Jianbing Shen.
\newblock Active visual information gathering for vision-language navigation.
\newblock In {\em ECCV}, 2020.

\bibitem{wijmans2019dd}
Erik Wijmans, Abhishek Kadian, Ari Morcos, Stefan Lee, Irfan Essa, Devi Parikh, Manolis Savva, and Dhruv Batra.
\newblock Dd-ppo: Learning near-perfect pointgoal navigators from 2.5 billion frames.
\newblock In {\em ICLR}, 2019.

\bibitem{xing2022selfmatch}
Huanlai Xing, Zhiwen Xiao, Dawei Zhan, Shouxi Luo, Penglin Dai, and Ke Li.
\newblock Selfmatch: Robust semisupervised time-series classification with self-distillation.
\newblock {\em International Journal of Intelligent Systems}, 2022.

\bibitem{ye2021auxiliary}
Joel Ye, Dhruv Batra, Erik Wijmans, and Abhishek Das.
\newblock Auxiliary tasks speed up learning point goal navigation.
\newblock In {\em Conference on Robot Learning}, pages 498--516. PMLR, 2021.

\bibitem{YinfengICLR2022saavn}
Yinfeng Yu, Wenbing Huang, Fuchun Sun, Changan Chen, Yikai Wang, and Xiaohong Liu.
\newblock Sound adversarial audio-visual navigation.
\newblock In {\em ICLR}, 2022.

\bibitem{zhao2022decoupled}
Borui Zhao, Quan Cui, Renjie Song, Yiyu Qiu, and Jiajun Liang.
\newblock Decoupled knowledge distillation.
\newblock In {\em CVPR}, pages 11953--11962, 2022.

\bibitem{zhao2022target}
Yusheng Zhao, Jinyu Chen, Chen Gao, Wenguan Wang, Lirong Yang, Haibing Ren, Huaxia Xia, and Si Liu.
\newblock Target-driven structured transformer planner for vision-language navigation.
\newblock In {\em ACM MM}, 2022.

\bibitem{zhu2021diagnosing}
Wanrong Zhu, Yuankai Qi, Pradyumna Narayana, Kazoo Sone, Sugato Basu, Xin~Eric Wang, Qi Wu, Miguel Eckstein, and William~Yang Wang.
\newblock Diagnosing vision-and-language navigation: What really matters.
\newblock {\em arXiv preprint arXiv:2103.16561}, 2021.

\bibitem{ObjectGoal01}
Yuke Zhu, Daniel Gordon, Eric Kolve, Dieter Fox, Li Fei-Fei, Abhinav Gupta, Roozbeh Mottaghi, and Ali Farhadi.
\newblock Visual semantic planning using deep successor representations.
\newblock In {\em ICCV}, 2017.

\end{thebibliography}
